\documentclass[a4paper]{llncs}

\usepackage{times}
\usepackage{graphicx}
\usepackage{amsmath}
\usepackage{amssymb}
\usepackage{bm}
\usepackage{enumerate}
\usepackage[pagebackref=false,breaklinks=true,colorlinks,bookmarks=false,linkcolor=blue,citecolor=blue,pdftitle={Joint Recognition and Segmentation of Actions via Probabilistic Integration of Spatio-Temporal Fisher Vectors},pdfauthor={Johanna Carvajal, Chris McCool, Brian Lovell, Conrad Sanderson}]{hyperref}

\def\Vec#1{{\boldsymbol{#1}}}

\begin{document}

\title
  {
  Joint Recognition and Segmentation of Actions via Probabilistic Integration of Spatio-Temporal Fisher Vectors
  }

\author
  {
  Johanna Carvajal,
  Chris McCool,
  Brian Lovell,
  Conrad Sanderson
  }

\institute
  {
  University of Queensland, Brisbane, Australia\\
  Queensland University of Technology, Brisbane, Australia\\
  NICTA, Australia\\
  Data61, CSIRO, Australia\\
  \textcolor{white}{\thanks{Published in: {\bf Lecture Notes in Computer Science (LNCS), Vol. 9794, pp.~115-127, 2016.}}}
  }

\maketitle

\begin{abstract}

\vspace{-6ex}
\begin{sloppypar}
We propose a hierarchical approach to multi-action recognition that performs joint classification and segmentation.
A~given video (containing several consecutive actions) is processed via a sequence of overlapping temporal windows.
Each frame in a temporal window is represented through selective low-level spatio-temporal features
which efficiently capture relevant local dynamics.
Features from each window are represented as a Fisher vector,
which captures first and second order statistics.
Instead of directly classifying each Fisher vector,
it is converted into a vector of class probabilities.
The final classification decision for each frame
is then obtained by integrating the class probabilities at the frame level,
which exploits the overlapping of the temporal windows.
Experiments were performed on two datasets: s-KTH (a~stitched version of the KTH dataset to simulate multi-actions),
and the challenging \mbox{CMU-MMAC} dataset.
On \mbox{s-KTH}, the proposed approach achieves an accuracy of 85.0\%,
significantly outperforming two recent approaches based on GMMs and HMMs which obtained 78.3\% and 71.2\%, respectively.
On \mbox{CMU-MMAC}, the proposed approach achieves an accuracy of 40.9\%,
outperforming the GMM and HMM approaches which obtained 33.7\% and 38.4\%, respectively.
Furthermore, the proposed system is on average 40 times faster than the GMM based approach. 

\end{sloppypar}
\end{abstract}

\vspace{-2ex}
\section{Introduction}
\vspace{-1ex}

Research on human action recognition can be divided into two areas:
{\bf (i)}~single-action recognition, and {\bf (ii)}~multi-action recognition.
In most computer vision literature, action recognition approaches have concentrated on single actions,
where each video to be classified contains only one action.
However, when observing realistic human behaviour in natural settings,
the fundamental problem is segmenting {and} recognising actions from a {\it multi-action} sequence~\cite{Buchsbaum2011}.
It~is challenging due to the high variability of appearances, shapes, possible occlusions,
large variability in the temporal scale and periodicity of human actions, the complexity of articulated motion,
the exponential nature of all possible movement combinations, as well as the prevalence of irrelevant background~\cite{Hoai2011,Qinfeng2008}.

Hoai et al.~\cite{Hoai2011} address joint segmentation and classification by classifying temporal regions using a multi-class Support Vector Machine (SVM) and performing segmentation using dynamic programming. 
A similar approach is presented in~\cite{YuCheng2014}, where the temporal relationship between actions is considered.
Borzeshi et al.~\cite{Borzeshi2013} proposed the use of hidden Markov models (HMMs) with irregular observations (termed HMM-MIO) to perform multi-action recognition. 
More recently, Carvajal et al.~\cite{Joha2014b} proposed to model each action using a Gaussian mixture model (GMM) approach where classification decisions are made using overlapping temporal windows.
A drawback of~\cite{Borzeshi2013,Hoai2011,YuCheng2014} is that they have a large number of parameters to optimise. 
Furthermore, \cite{Borzeshi2013} requires an extra stage to reduce dimensionality due to use of very high dimensional feature vectors,
while~\cite{Hoai2011,YuCheng2014} require fully labelled annotations for training.
One downside of~\cite{Joha2014b} is that for each action and scenario, a~model with a large number of Gaussians is required,
making the system computationally expensive.

Typically, the aforementioned approaches used for the multi-action recognition task  can be classified as either generative or discriminative models. 
The approaches presented in~\cite{Borzeshi2013,Joha2014b} are generative models, while those presented in~\cite{YuCheng2014,Hoai2011} are discriminative models.
Generative and discriminative models have complementary strengths.
Generative models can easily deal with variable length sequences and missing data,
while also being easier to design and implement~\cite{Jaakkola98,Lasserre2007}.
In contrast, discriminative models often achieve superior classification and generalisation performance~\cite{Jaakkola98,Lasserre2007}.
An ideal recognition system would hence combine these two separate but complementary approaches.

The Fisher vector (FV) approach~\cite{Jaakkola98,Perronnin2011,JorgeSanchez2013} allows for the fusion of both generative and discriminative models.
In contrast to the popular Bag of Words (BoW) approach~\cite{HengWang2009} which describes images by histograms of visual words,
the FV approach describes images by deviations from a probabilistic visual vocabulary model.
The resulting vectors can then be used by an SVM for final classification.
Recently, FV has been successfully applied to the single-action recognition problem~\cite{Oneata2013,HengWang2013}.

\begin{figure}[!b]
  \centering
  \begin{minipage}{1\textwidth}
  \centering
  \begin{minipage}{0.54\textwidth}
    \includegraphics[width=1\columnwidth]{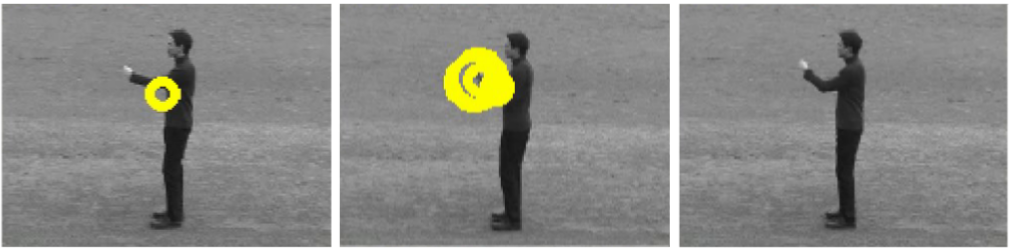}\\
    \includegraphics[width=1\columnwidth]{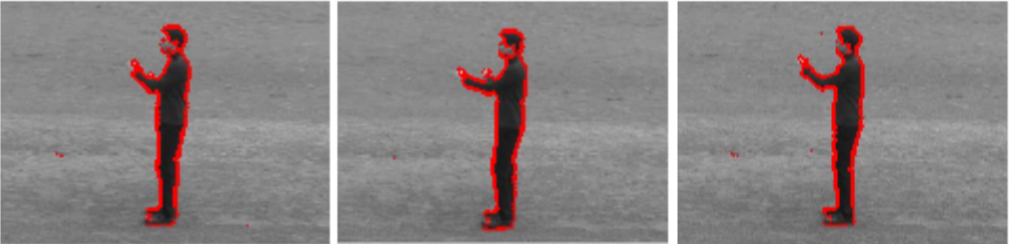}
  \end{minipage}
  \hfill
  \begin{minipage}{0.45\textwidth}
  \caption
    {
    Top row: feature extraction based on Spatio-Temporal Interest Points (STIPs) is often unstable, imprecise and overly sparse.
    Bottom row: interest pixels (marked in red) obtained using magnitude of gradient.
    }
  \vspace{5ex}
  \end{minipage}
  \end{minipage}
\label{fig:Descriptors}
\end{figure}

A reliable low-level feature descriptor is a crucial stage for the success of an action recognition system.
One popular descriptor for action recognition is Spatio-Temporal Interest Points (STIPs)~\cite{Laptev2005}.
However, STIP based descriptors have several drawbacks \cite{Liangliang2010,Kliper2012}:
they are computationally expensive, unstable, imprecise and can result in unnecessarily sparse detections.  
See Fig.~\ref{fig:Descriptors} for a demonstration of STIP based detection. 
Other feature extraction techniques used for action recognition include gradients~\cite{Joha2014b} and optical flow~\cite{Ali2010,Kliper2012}.
Each pixel in the gradient image helps extract relevant information, eg.~edges (see Fig.~\ref{fig:Descriptors}).
Since the task of action recognition is based on a sequence of frames, optical flow provides an efficient way of capturing the local dynamics~\cite{Kliper2012}.

\textbf{Contributions.}
In this paper we propose a novel hierarchical system to perform multi-action segmentation and recognition.
A given video is processed as a sequence of overlapping temporal windows.
Each frame in a temporal window is represented through selective low-level spatio-temporal features,
based on a combination of gradients with optical flow.
Interesting features from each temporal window are then pooled and processed to form a Fisher vector.
Instead of directly classifying each Fisher vector,
a multi-class SVM is employed to generate a vector of probabilities, with one probability per class.
The final classification decision (action label) for each frame
is then obtained by integrating the class probabilities at the frame level,
which exploits the overlapping of the temporal windows.
The proposed system hence combines the benefits of generative and discriminative models.

To the best of our knowledge, the combination of probabilistic integration with Fisher vectors
is novel for the {\it multi-action} segmentation and recognition problem. 
In contrast to~\cite{Borzeshi2013,Hoai2011,YuCheng2014},
the proposed system requires fewer parameters to be optimised.
We also avoid the need for a custom dynamic programming definition as in~\cite{Hoai2011,YuCheng2014}.
Lastly, unlike the GMM approach proposed in~\cite{Joha2014b}
the proposed method requires only one GMM for all actions,
making it considerably more efficient.

The paper is continued as follows.
An overview of prior work on feature descriptors as well as single- and multi-action recognition is in Section~\ref{sec:related}.
We then describe the proposed method in Section~\ref{sec:proposed_method}.
A comparative evaluation against previous multi-action recognition methods is presented in Section~\ref{sec:experiments}.
The main findings and potential areas for future work are given in Section~\ref{sec:conclusions}.

\section{Related Work}
\label{sec:related}
\vspace{-1ex}

In this section, we present an overview of notable methods for action recognition.
We first describe several popular descriptors used for action recognition.
This is followed by an overview of techniques for single-action recognition.
We then present a summary of approaches for multi-action segmentation and recognition.

\vspace{-0.5ex}
\subsection{Low-Level Descriptors}
\label{subsec:descriptors}
\vspace{-0.5ex}

Several descriptors have been used to represent human actions. Among them, we can find descriptors based on Spatio-Temporal Interest Points (STIPs)~\cite{Laptev2005}, gradients~\cite{Joha2014b}, and optical flow~\cite{Ali2010}.
The STIP based detector finds interest points and is a space-time extension of the Harris and F\"{o}rstner interest point detector~\cite{Fostner87,Harris88}. STIP descriptors have led to relatively good action recognition performance~\cite{Borzeshi2012}. However, STIP based descriptors have several drawbacks: 
{\bf (i)}~they focus on local spatio-temporal information instead of global motion,
{\bf (ii)}~they can be unstable and imprecise (varying number of STIP detections) leading to low repeatability,
{\bf (iii)}~they are computationally expensive,
{\bf (iv)}~produce sparse detections,
and
{\bf (v)}~are high dimensional with default descriptors of 162 dimensions.

The Scale Invariant Feature Transform (SIFT) was first described in~\cite{Lowe1999}. 
Key locations are chosen at maxima and minima of a difference of Gaussian function applied in scale space.
Orientation is assigned to each key point to achieve invariance to image rotation.
Finally, the key-point descriptor is created taking a neighbourhood around the key point.
SIFT features holds  positive attributes such as local invariance to transformation, rotation, and scale.
However, its applicability to large datasets or real-time applications is limited due to it its relatively high dimensionality and high computational demands~\cite{Elgayar2013}.

A video feature descriptor using dense trajectories and motion boundary descriptors is proposed in~\cite{HengWang2013}.
The dense representation allows for good coverage of foreground motion as well as of the surrounding context.
This descriptor has become popular for the single-action recognition task
despite of being relatively computationally expensive and high dimensional, with the final number of dimensions being 426.

Gradients have been used as a relatively simple yet effective video representation~\cite{Joha2014b}.
Each pixel in the gradient image helps extract relevant information, eg.~edges of a subject.
Gradients can be computed at every spatio-temporal location $(x, y, t)$ in any direction in a video.
Lastly,
since the task of action recognition is based on an ordered sequence of frames,
optical flow can be used to provide an efficient way of capturing local dynamics and motion patterns in a scene~\cite{Kliper2012}.

\vspace{-0.5ex}
\subsection{Single-Action Recognition}
\vspace{-0.5ex}

Hidden Markov models (HMMs) have been used in conjunction with shape-context features to recognise single actions~\cite{Mendoza2007}. 
The  shape-context features are extracted using the image contours and dividing the region where the person is into uniform tiles.
For each tile a feature vector is generated.
The discrete cosine transform is used in order to minimise redundancy and to compress the data.
Continuous HMM with mixed Gaussian output probability is employed with a simple left to right topology.
 
Gaussian Mixture Models (GMMs) have also been explored for the single-action detection and classification.
For the approach presented in~\cite{WeiyaoLin2008}, each action is represented by a combination of GMMs.
Each action is modelled by two sets of feature attributes.
The first set represents the change of body size, while the second represents the speed of the action. 
Features with high correlations for describing actions are grouped into the same Category Feature Vector (CFV).
All CFVs related to the same category are then modelled using a GMM.
A Confident-Frame-based Recognising algorithm is used for recognition,
where the video frames which have high confidence are used as a specialised model for classifying the rest of the video frames.

A video sequence can be also represented as a Bag of Words (BoW)~\cite{HengWang2009}. 
This approach and its variants are among the most popular approaches for action recognition~\cite{HengWang2013,HengWang2009}. 
The standard approach consists of four main steps: low-level feature extraction, offline codebook generation, feature encoding and pooling, and normalisation. 
The offline codebook, also known as visual dictionary, is constructed using {\it k}-means clustering of a large training dataset.  
Visual words are then defined as the centers of the clusters,
with the size of visual dictionary equal to the number of the clusters. 
To encode low-level features, each feature vector is assigned to its closest dictionary word using Euclidean distance.
A given video is then represented as a frequency histogram over the visual words. 
Normalised histograms can then be used by linear classifiers.

The Fisher vector (FV) representation can be thought as an evolution of the BoW representation, encoding additional information~\cite{Perronnin2011,HengWang2013}. 
Rather than encoding the frequency of the descriptors, FV encodes the deviations from a probabilistic version of the visual dictionary (which is typically a GMM).
This is done by computing the gradient of the sample log-likelihood with respect the parameters of the dictionary model. 
Since more information is extracted, a smaller visual dictionary size can be used than for BoW, in order to achieve the same or better performance. %~\cite{JorgeSanchez2013}.
Finally, $L_2$ and power normalisation is often used to improve performance in combination with linear classifiers~\cite{JorgeSanchez2013}.

FV has been used for a wide range of applications in the computer vision community, such as image retrieval~\cite{Douze2011_cvpr},
as well as image representation and classification~\cite{Perronnin2011,JorgeSanchez2013}.
Recently, FV has been also successfully applied to the single-action recognition problem~\cite{Oneata2013,HengWang2013}. 
Descriptors based on motion boundary histograms and Scale-Invariant Feature Transform (SIFT) are used in conjunction with FV in~\cite{Oneata2013}.
The dimensionality of each descriptor is reduced to $64$ dimensions via Principal Component Analysis (PCA). 
A similar approach is presented in~\cite{HengWang2013},
where the SIFT descriptors are replaced by histograms of gradients, optical flow, and trajectories.  
To combine various descriptor types, FVs derived from each descriptor type are concatenated.

Note that both the FV and BoW approaches build on top of low-level features, which typically describe small image areas.
The low-level features may or may not have robustness to scale \& rotations (see Section~\ref{subsec:descriptors}).
However, even if the low-level features are not directly robust, it has been previously shown that the BoW approach (and by extension the FV approach)
can still exhibit robustness to moderate variations~\cite{Cardinaux_IEEE_TSP_2006,Sanderson_PR_2006,Sanderson_LNCS_2009}.
The reasoning is summarised as follows.
When an object undergoes moderate scale changes or rotations, the dominant effect on many small parts is translation.
Both FV and BoW do not have rigid constraints on the location of object parts,
and also do not have rigid constraints on spatial relations between parts.
This lack of constraints allows for movement (translation) and dispersion of small image areas,
which in turn leads to a degree of overall robustness of FV and BoW.

The action recognition problem can be also solved by decomposing actions into sub-actions or atoms~\cite{LiminWang_2013,LiminWang_2014}. 
In the former, motion atoms are obtained using a discriminative clustering method.  
These atoms are basic units used to construct motion phrase with a longer scale. 
To this end, a bottom-up phrase construction algorithm and a greedy selection method are used.
Motion phrase is composed of multiple atomic motion units.
The latter work presents a latent hierarchical model. 
This hierarchical model has a tree structure, where each node represents a sub-action.
Two latent variables are used to represent the starting and ending time points of each sub-action.

\vspace{-0.5ex}
\subsection{Multi-Action Recognition}
\label{subsec:prev_multi_action_recog}
\vspace{-0.5ex}

In contrast to single action recognition, relatively little work exists on multi-action recognition.
Multi-action recognition, in our context, consists of segmenting and recognising separate actions from an image sequence, where one person performs a sequence of such actions~\cite{Qinfeng2008}.
The process for segmenting and recognising multiple actions in a video can be solved either as two independent problems or as a joint problem.

Two methods~\cite{Borzeshi2013,Hoai2011} have been applied to realistic multi-action datasets.
Hoai et al.~\cite{Hoai2011} deal with the dual problem of human action segmentation and classification.
The recognition model is trained using labelled data with a multi-class SVM.
Once the model for all actions has been obtained, the video segmentation and recognition is done using dynamic programming, maximising the SVM score of the winning class while suppressing those of the non-maximum classes.
The feature mapping depends on the dataset employed, and includes trajectories, features extracted from binary masks and STIPs.

Although the method proposed in~\cite{Hoai2011} is promising, it has several drawbacks. One drawback is the requirement of fully labelled annotations for training.
Furthermore, it suffers from the limitations of dynamic programming where writing the code that evaluates sub-problems in the most efficient order is often nontrivial~\cite{Wagner1995}.
Also, the binary masks are not always available and the STIP descriptors have deficiencies as mentioned in Section~\ref{subsec:descriptors}.
This method also requires an extensive search for optimal parameters.

An approach termed Hidden Markov Model for Multiple, Irregular Observations (HMM-MIO)~\cite{Borzeshi2013} has also been proposed for the multi-action recognition task.
HMM-MIO jointly segments and classifies observations which are irregular in time and space, and are characterised by high dimensionality.
The high dimensionality is reduced by probabilistic PCA. 
Moreover, HMM-MIO deals with heavy tails and outliers exhibited by empirical distributions by modelling the observation densities with the long-tailed Student's~$t$ distribution.
HMM-MIO requires the search of the following four optimal parameters:
{\bf (i)}~the resulting reduced dimension,
{\bf (ii)}~the number of components in each observation mixture,
{\bf (iii)}~the degree of the $t$-distribution,
and
{\bf (iv)}~the number of cells (or regions) used to deal with space irregularity.  
As feature descriptors, HMM-MIO extracts STIPs, with the default $162$-dimensional descriptor.
HMM-MIO hence suffers from the drawback of a large search of optimal parameters and the use of STIP descriptors.

A recent approach to multi-action recognition via stochastic modelling of optical flow and gradients is presented in~\cite{Joha2014b}.
Each action is modelled using a GMM.
Action segmentation is achieved by integrating classification decisions on overlapping temporal windows.
In each window, the average log-likelihood is obtained for each model using a large number of components (1024).
This approach outperforms HMM-MIO, but is limited by being a purely generative model.
Moreover, due to the large number of components per action model,
the calculation of the likelihoods using all action models is computationally very expensive.

\newpage
\section{Proposed Method}
\label{sec:proposed_method}

\noindent
The proposed system has a hierarchical nature, stemming from progressive reduction and transformation of information, starting at the pixel level.
The system is comprised of four main components:

\begin{enumerate}[{\bf ~(i)}]

\item
Division of a given video into overlapping multi-frame temporal windows,
followed by extracting interesting low-level spatio-temporal features from each frame in each window.

\item
Pooling of the interesting features from each temporal window to generate a sequence of Fisher vectors.

\item
Conversion of each Fisher vector into a vector of class probabilities with the aid of a multi-class SVM.

\item
Integration of the class probabilities at the frame level,
leading to the final classification decision (action label) for each frame.

\end{enumerate}

\noindent
Each of the components is explained in more detail in the following subsections.

\subsection{Overlapping and Selective Feature Extraction}
\label{subsec:features}

A video ${\mathcal{V}} = \left( {\bm{I}}_t \right)_{t=1}^T$ is an ordered set of $T$ frames.
We divide ${\mathcal{V}}$ into a set of overlapping temporal windows $\left( \mathcal{W}_s \right)_{s=1}^{S}$,
with each window having a length of $L$ frames.
To~achieve overlapping, the start of each window is one frame after the start of the preceding window.
Each temporal window is hence defined as a set of frame identifiers:
\mbox{$\mathcal{W}_s = \left( t_{\mathrm{start}},~ \ldots,~ t_{\mathrm{start}-1+L}\right)$}. 

Each frame $\bm{I}_t \in \mathbb{R}^ {r\times c}$ can be represented by a set of feature vectors $F_t = \{\bm{f}_p\}_{p=1}^{N_t}$
(with $N_t < r \cdot c$) corresponding to interesting pixels.
Following~\cite{Joha2014b}, we first extract the following $D=14$ dimensional feature vector for each pixel in a given frame~$t$:
\begin{equation}\label{eq:features}
\bm{f} = \left[ \; x, \; y, \; \bm{g}, \; \bm{o} \; \right]
\end{equation}

\noindent
where $x$ and $y$ are the pixel coordinates, while $\bm{g}$ and $\bm{o}$ are defined as:
\begin{small}
\begin{eqnarray}
\hspace{-2ex} \bm{g} \hspace{-1ex} & = & \hspace{-1ex}
\left[ \; |J_x|, \; |J_y|, \; |J_{yy}|, \; |J_{xx}|, \; \sqrt{J_x^2 + J_y^2}, \; \text{atan} \frac{|J_y|}{|J_x|} \; \right]
\label{eq:features2}\\
\hspace{-2ex} \bm{o} \hspace{-1ex} & = & \hspace{-1ex}
\left[\; u, \;\;  v, \;\; \frac{\partial{u}}{\partial{t}}, \;\;  
\frac{\partial{v}}{\partial{t}}, \;\; 
\left (\frac{\partial u}{\partial x} + \frac{\partial v}{\partial y} \right ), \;\;
\left (\frac{\partial v}{\partial x} - \frac{\partial u}{\partial y} \right ) \;
\right]
\label{eq:features3}
\end{eqnarray}%
\end{small}%

The first four gradient-based features in Eq.~(\ref{eq:features2}) represent the first and second order intensity gradients at pixel location $(x,y)$.
The last two gradient features represent gradient magnitude and gradient orientation.
The optical flow based features in Eq.~(\ref{eq:features3}) represent in order:
the horizontal and vertical components of the flow vector,
the first order derivatives with respect to time,
the divergence and vorticity of optical flow~\cite{Ali2010}.

Typically only a subset of the pixels in a frame correspond to the object of interest.
As such, we are only interested in pixels with a gradient magnitude greater than a threshold~$\tau$~\cite{KaiGuo2013}.
We discard feature vectors from locations with a small magnitude.
In other words, only feature vectors corresponding to interesting pixels are kept.
This typically results in a variable number of feature vectors per frame.
See the bottom part in Fig.~\ref{fig:Descriptors} for an example of the retained pixels.

\subsection{Representing Windows as Fisher Vectors}
\label{subsec:train}

Given a set of feature vectors, the Fisher Vector approach encodes the deviations from a probabilistic visual dictionary,
which is typically a diagonal GMM.
The parameters of a GMM with $K$ components can be expressed as $\{w_k,\bm{\mu}_k,\bm{\sigma}_k\}_{k=1}^{K}$,
where, $w_k$ is the weight, $\bm{\mu}_k$ is the mean vector, and $\bm{\sigma}_k$ is the diagonal covariance matrix
for the $k$-th Gaussian.
The parameters are learned using the Expectation Maximisation algorithm~\cite{Bishop_PRML_2006} on training data.

For each temporal window $\mathcal{W}_s$, the feature vectors are pooled into set $X$
containing \mbox{$N = \sum_{t \in \mathcal{W}_s} N_t$} vectors.
The deviations from the GMM are then accumulated using~\cite{JorgeSanchez2013}:
\begin{eqnarray}
\mathcal{G}_{\Vec{\mu}_{k}}^{X}    & = & \frac{1}{N\sqrt{w_k}} \sum\nolimits_{n=1}^{N} \gamma_n(k)\left( \frac{\bm{f}_n - \bm{\mu}_k}{\bm{\sigma}_k} \right)\\
\mathcal{G}_{\Vec{\sigma}_{k}}^{X} & = & \frac{1}{N\sqrt{2w_k}} \sum\nolimits_{n=1}^{N} \gamma_n(k)\left[ \frac{\left(\bm{f}_n - \bm{\mu_k}\right)^2}{\bm{\sigma}_k^2} -1 \right]
\end{eqnarray}%

\noindent where vector division indicates element-wise division
and $\gamma_n(k)$ is the posterior probability of $\bm{f}_n$ for the $k$-th component:
\begin{equation}
\gamma_n(k) = \frac{w_k\mathcal{N}(\bm{f}_n|\bm{\mu}_k, \bm{\sigma}_k)}{\sum\nolimits_{i=1}^{K}w_i\mathcal{N}(\bm{f}_n|\bm{\mu}_i, \bm{\sigma}_i)}
\end{equation}%

\noindent
The Fisher vector for window $\mathcal{W}_s$ is represented as the concatenation of
{\small $\mathcal{G}_{\Vec{\mu}_{k}}^{X}$} and {\small $\mathcal{G}_{\Vec{\sigma}_{k}}^{X}$} (for {\small $k$~=~$1, \ldots, K$}) into vector~$\Vec{\Phi}_s$.
As {\small $\mathcal{G}_{\Vec{\mu}_{k}}^{X}$} and {\small $\mathcal{G}_{\Vec{\sigma}_{k}}^{X}$} are \mbox{$D$-dimensional},
$\Vec{\Phi}_s$ has the dimensionality of $2DK$.
Note that we have omitted the deviations for the weights as they add little information~\cite{JorgeSanchez2013}.

\subsection{Generation of Probability Vectors}

For each Fisher vector we generate a vector of probabilities, with one probability per action class.
First, a multi-class SVM~\cite{Crammer_2001} is used to predict class labels, outputting a set of raw scores.
The scores are then transformed into a probability distribution over classes by applying Platt scaling~\cite{Platt99}.
The final probability vector derived from Fisher vector $\Vec{\Phi}_s$ is expressed as:
\begin{equation}
\Vec{q}_s = \left[ ~ P(l=1|\Vec{\Phi}_s), ~ \cdots, ~ P(l=A | \Vec{\Phi}_s) ~ \right]
\end{equation}%

\noindent
where $l$ indicates an action class label and $A$ is the number of action classes.
The parameters for the multi-class SVM are learned using Fisher vectors obtained from pre-segmented actions in training data.

\subsection{Integrating Probability Vectors to Label Frames}
\label{subsec:multiaction}

As the temporal windows are overlapping, each frame is present in several temporal windows.
We exploit the overlapping to integrate the class probabilities at the frame level.
The total contribution of the probability vectors to each frame $t$ is calculated by:
\begin{equation}
\Vec{Q}_t = \sum\nolimits_{s=1}^{S} \mathbf{1}_{\mathcal{W}_s} (t) \cdot \Vec{q}_s
\end{equation}

\noindent
where $\mathbf{1}_{\mathcal{W}_s} (t)$ is an indicator function, resulting in 1 if \mbox{$t \in \mathcal{W}_s$}, and 0 otherwise.
The estimated action label for frame $t$ is then calculated as:
\begin{equation}
\widehat{l}_t = \underset{l=1,\ldots,A} {\mathrm{arg~max}} ~ \Vec{Q}_t^{[l]}
\vspace{-1ex}
\end{equation}%
\noindent
where $\Vec{Q}_t^{[l]}$ indicates the $l$-th element of $\Vec{Q}_t$.

\section{Experiments}
\label{sec:experiments}

We evaluated our proposed method for joint action segmentation and recognition on two datasets:
{\bf (i)} a stitched version of the KTH dataset~\cite{Schuldt2004},
and
{\bf (ii)} the challenging Carnegie Mellon University Multi-Modal Activity Dataset (CMU-MMAC)~\cite{DeLaTorreHMV09}.
The results are reported in terms of frame-level accuracy as the ratio between the number of matched frames over the total number of frames.

\subsection{Datasets}

The \textbf{s-KTH} (stitched KTH) dataset is obtained by simply concatenating existing single-action instances into sequences~\cite{Borzeshi2013}. 
The KTH dataset contains 25 subjects performing 6 types of human actions and 4 scenarios. The scenarios vary from indoor, outdoor, scale variations, and different clothes.
Each original video of the KTH dataset~\cite{Schuldt2004} contains an individual performing the same action. 
This action is performed four times and each subdivision or action-instance (in terms of start-frame and end-frame) is provided as part of the dataset.
This dataset contains 2391 action-instances, with a length between 1 and 14 seconds~\cite{Baccouche2011}.
The image size is 160$\times$120 pixels, and temporal resolution is 25 frames per second.

\begin{figure*}[tb!]
  \centering
  \begin{minipage}{1\textwidth}
    \begin{minipage}{1\textwidth}
      \centering
      \includegraphics[width=0.19\textwidth]{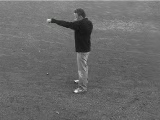}
      \includegraphics[width=0.19\textwidth]{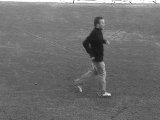}
      \includegraphics[width=0.19\textwidth]{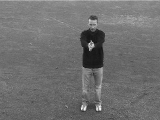}
      \includegraphics[width=0.19\textwidth]{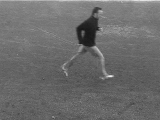}
      \includegraphics[width=0.19\textwidth]{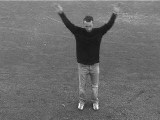}
    \end{minipage}
    \vspace{-2ex}    
    \caption{Example of a multi-action sequence in the stitched version of the KTH dataset (s-KTH): boxing, jogging, hand clapping, running, hand waving and walking.}
    \label{fig:stiched_example}
  \end{minipage}
  
  \vspace{2ex}
  
  \begin{minipage}{1\textwidth}
    \begin{minipage}{1\textwidth}
      \centering
      \includegraphics[width=0.2425\textwidth]{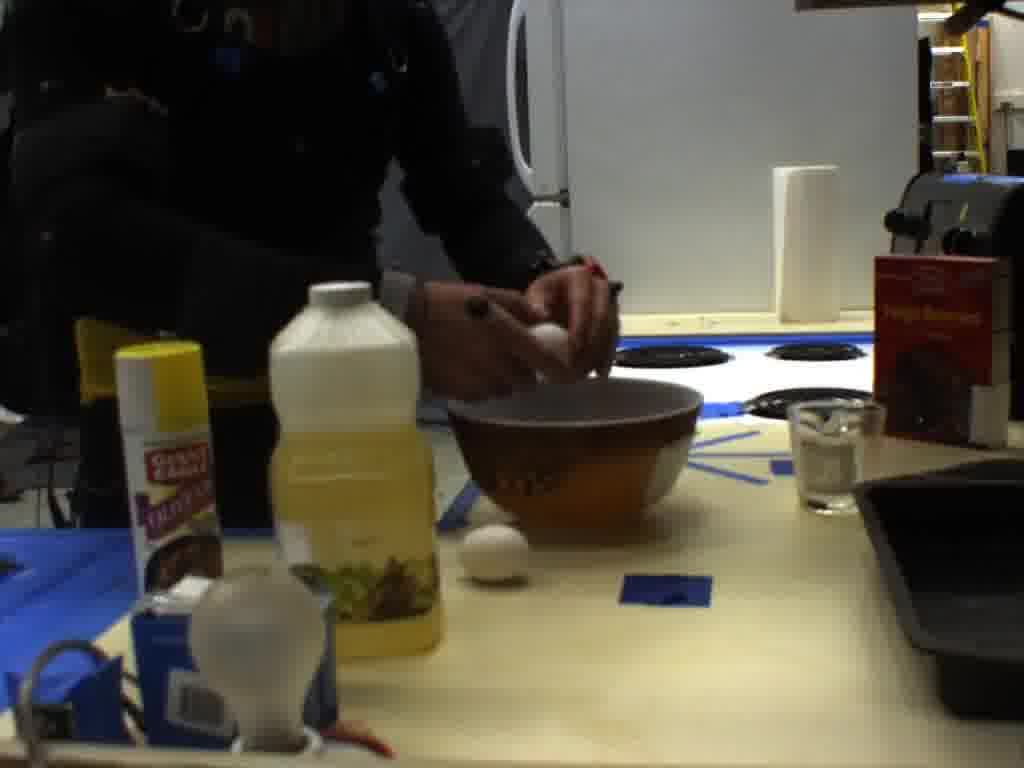}
      \includegraphics[width=0.2425\textwidth]{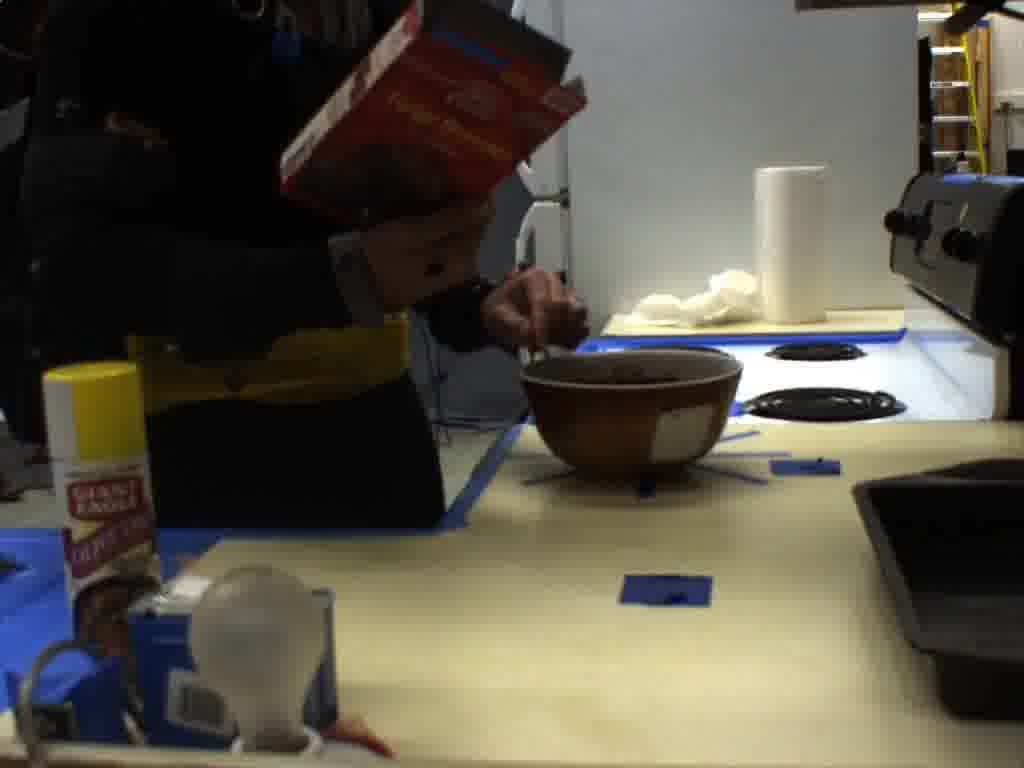}
      \includegraphics[width=0.2425\textwidth]{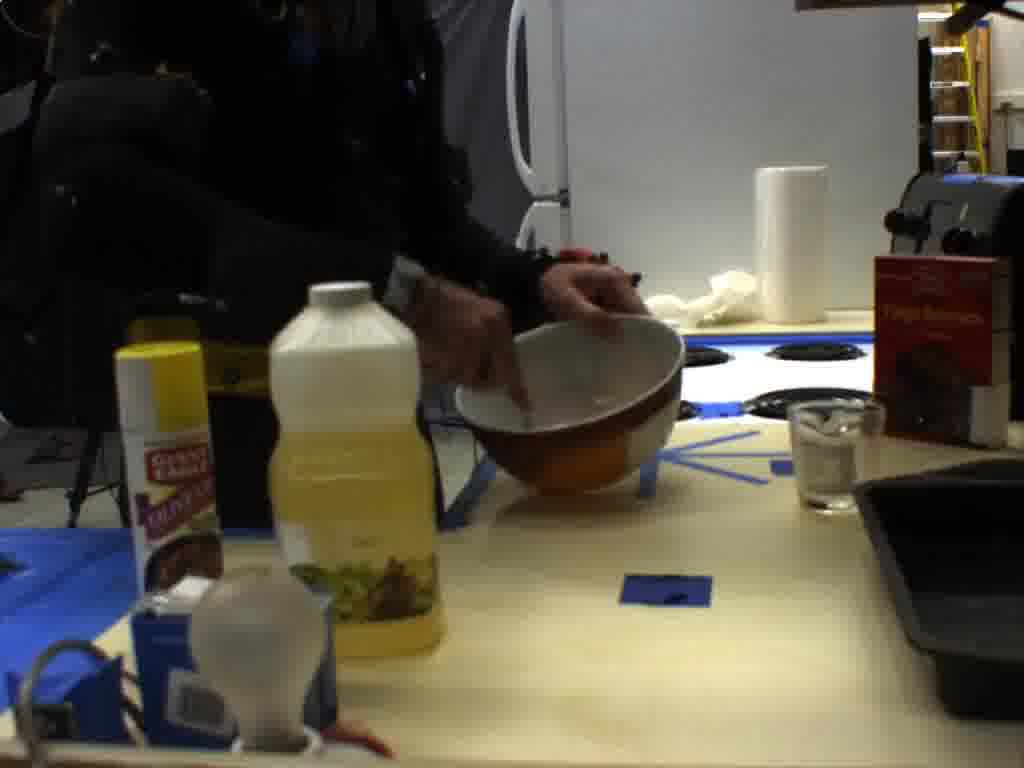}
      \includegraphics[width=0.2425\textwidth]{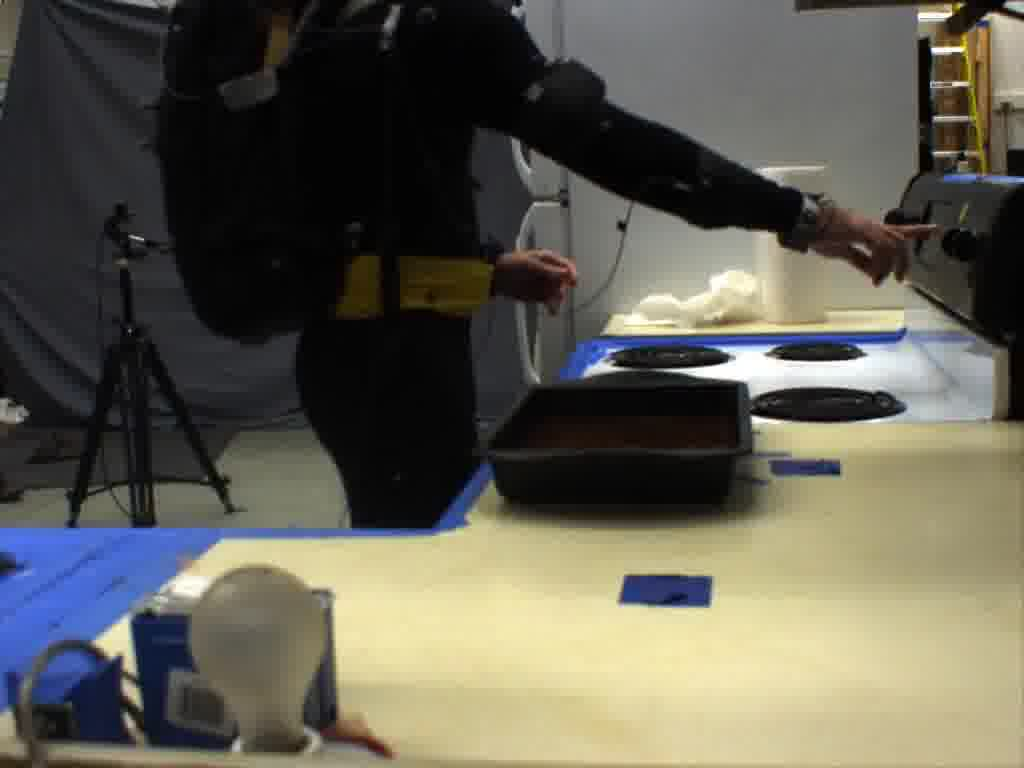}
    \end{minipage}
    \vspace{-2ex}    
    \caption{Example of a challenging multi-action sequence in the CMU-MMAC kitchen dataset: crack, read, stir, and switch-on.}
    \label{fig:cmu_example}
  \end{minipage}
  \vspace{-2ex}
\end{figure*}

The action-instances (each video contains four instances of the action) were picked randomly,
alternating between the two groups of \{boxing, hand-waving, hand-clapping\} and \{walking, jogging, running\} to accentuate action boundaries. 
See Fig.~\ref{fig:stiched_example} for an example.  
The dataset was divided into two sets as in~\cite{Borzeshi2013,Joha2014b}: one for training and one for testing.
In total, 64~and 36 multi-action videos were used for training and testing, respectively.
We used 3-fold cross-validation.

The \textbf{CMU-MMAC} dataset is considerably more challenging as it contains realistic multi-action videos~\cite{DeLaTorreHMV09}. 
A~kitchen was built to record subjects preparing and cooking food according to five recipes.
This dataset has occlusions, a cluttered background, and many distractors such as objects being deliberately moved.
For our experiments we have used the same subset as per~\cite{Borzeshi2013}, which contains 12 subjects making brownies. 
The subjects were asked to make brownies in a natural way (no instructions were given).
Each subject making the brownie is partially seen, as shown in Fig.~\ref{fig:cmu_example}.

The videos have a high resolution and are longer than in \mbox{s-KTH}.
The image size is  1024$\times$768 pixels, and temporal resolution is 30 frames per second.
The average duration of a video is approximately 15,000 frames
and the average length of an action instance is approximately 230 frames (7.7s), with a minimum length of 3 frames (0.1s) and a maximum length of 3,269 frames (108s)~\cite{Borzeshi2013}.
The dataset was annotated using 14 labels, including the actions {\it close}, {\it crack}, {\it open}, {\it pour}, {\it put}, {\it read}, {\it spray}, {\it stir}, {\it switch-on}, {\it take}, {\it twist-off}, {\it twist-on}, {\it walk},
and the remaining actions (eg.~frames in between two distinct actions) were grouped under the label {\it none}~\cite{Spriggs2009}.
We used 12-fold cross-validation, using one subject for testing on a rotating basis.
The experiments were implemented with the aid of the Armadillo C++ library~\cite{Armadillo2016}.

\subsection{Setup}

All videos were converted into gray-scale. Additionally, the videos from the CMU-MMAC dataset were re-scaled to 128$\times$96 to reduce computational requirements.
Based on~\cite{Joha2014b} and preliminary experiments on both datasets,
we used $\tau = 40$, where $\tau$ is the threshold used for selection of interesting low-level feature vectors (Section~\ref{subsec:features}) .
Although the interesting feature vectors are calculated in all frames, we only use the feature vectors extracted from every second frame in order to speed up processing.

Parameters for the visual vocabulary (GMM) were learned using a large set of descriptors obtained from training videos using the iterative Expectation-Maximisation algorithm~\cite{Bishop_PRML_2006}.
Specifically, we randomly sampled 100,000 feature vectors for each action and then pooled all the resultant feature vectors from all actions for training.
Experiments were performed with three separate GMMs with varying number of components: {$K=\{64,128,256\}$}. 
We have not evaluated larger values of $K$ due to increased computational complexity and hence the exorbitant amount of time required to process the large CMU-MMAC dataset.

To learn the parameters of the multi-class SVM, we used video segments containing single actions.
For s-KTH this process is straightforward as the videos have been previously segmented.
The CMU-MMAC dataset contains continuous multi-actions.
For this reason, to train our system we obtain one Fisher vector per action in each video,
using the low-level feature vectors belonging to that specific action.

\subsection{Effect of Window Length and Dictionary Size}
\label{sec:comparative_evaluation}
\vspace{-1ex}

\begin{figure}[!t]
\centering
\begin{minipage}{0.8\textwidth}
\centering
  \begin{minipage}{0.04\textwidth}
    \rotatebox{90}{\sf\footnotesize~~~accuracy (\%)}
  \end{minipage}
  \hfill
  \begin{minipage}{0.465\textwidth}
    \centering
    {\sf\footnotesize~~~s-KTH}\\
     \includegraphics[width=1.0\textwidth]{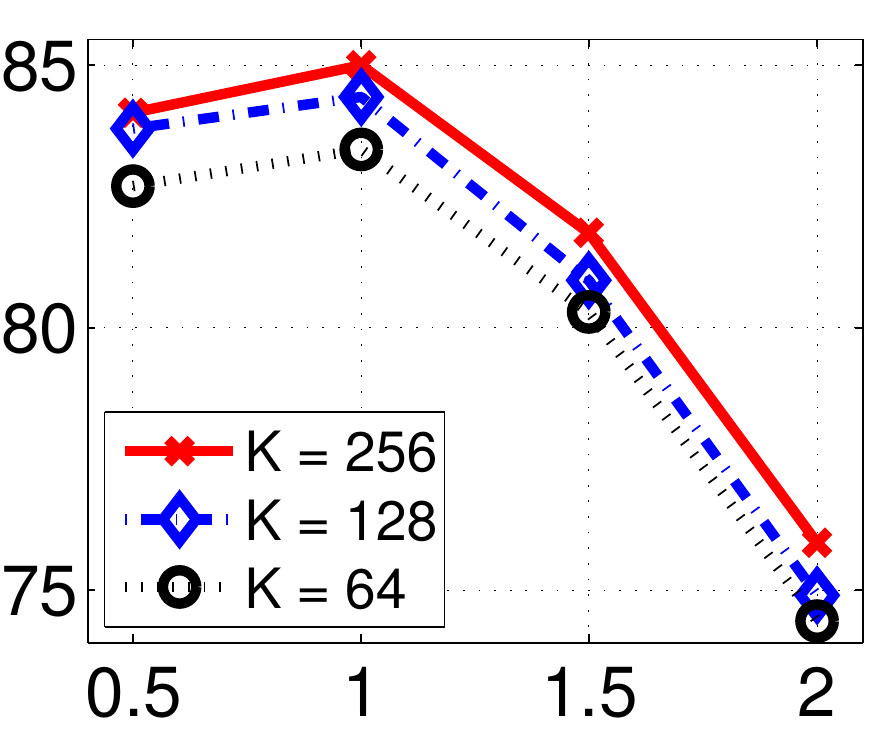}\\
    \vspace{-1ex}
    {\sf\footnotesize~~~segment length (sec)}
  \end{minipage}
  \hfill
  \begin{minipage}{0.465\textwidth}
    \centering
    {\sf\footnotesize~~~CMU-MMAC}\\
    \includegraphics[width=1.0\textwidth]{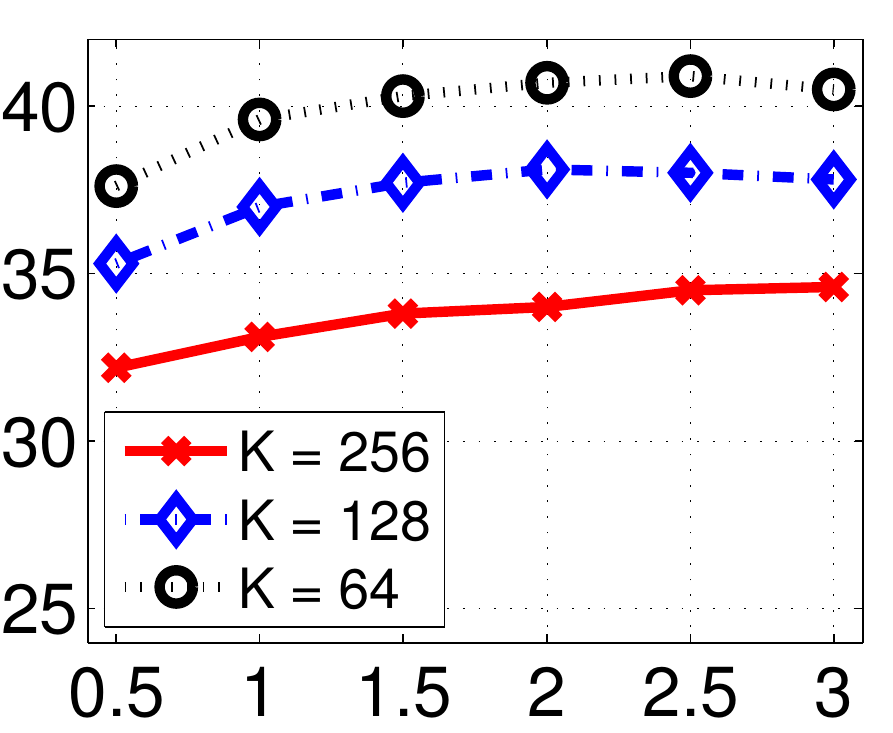}\\
    \vspace{-1ex}
    {\sf\footnotesize~~~segment length (sec)}
  \end{minipage}
\end{minipage}
\vspace{-1ex}
\caption{Performance of the proposed {\bf PI-FV} approach for varying the segment length on the \mbox{s-KTH} and \mbox{CMU-MMAC} datasets, in terms of average frame-level accuracy over the folds.}
\label{fig:Influence_L_FV}

\vspace{2ex}

\begin{minipage}{0.8\textwidth}
\centering
  \begin{minipage}{0.04\textwidth}
    \rotatebox{90}{\sf\footnotesize~~~accuracy (\%)}
  \end{minipage}
  \hfill
  \begin{minipage}{0.465\textwidth}
    \centering
    {\sf\footnotesize~~~s-KTH}\\
     \includegraphics[width=1.0\textwidth]{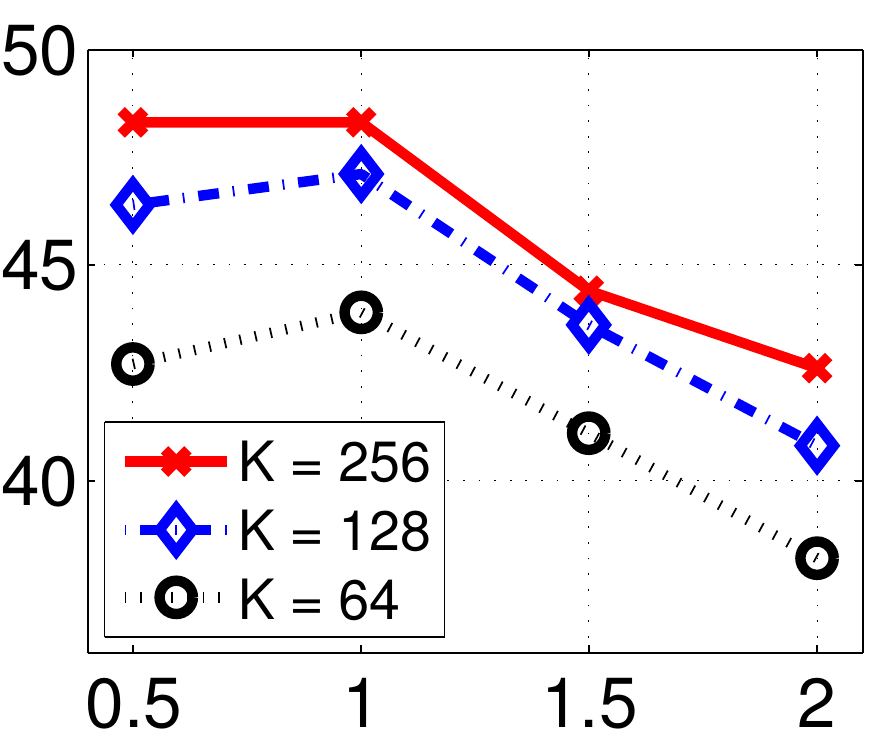}\\
    \vspace{-1ex}
    {\sf\footnotesize~~~segment length (sec)}
  \end{minipage}
  \hfill
  \begin{minipage}{0.465\textwidth}
    \centering
    {\sf\footnotesize~~~CMU-MMAC}\\
    \includegraphics[width=1.0\textwidth]{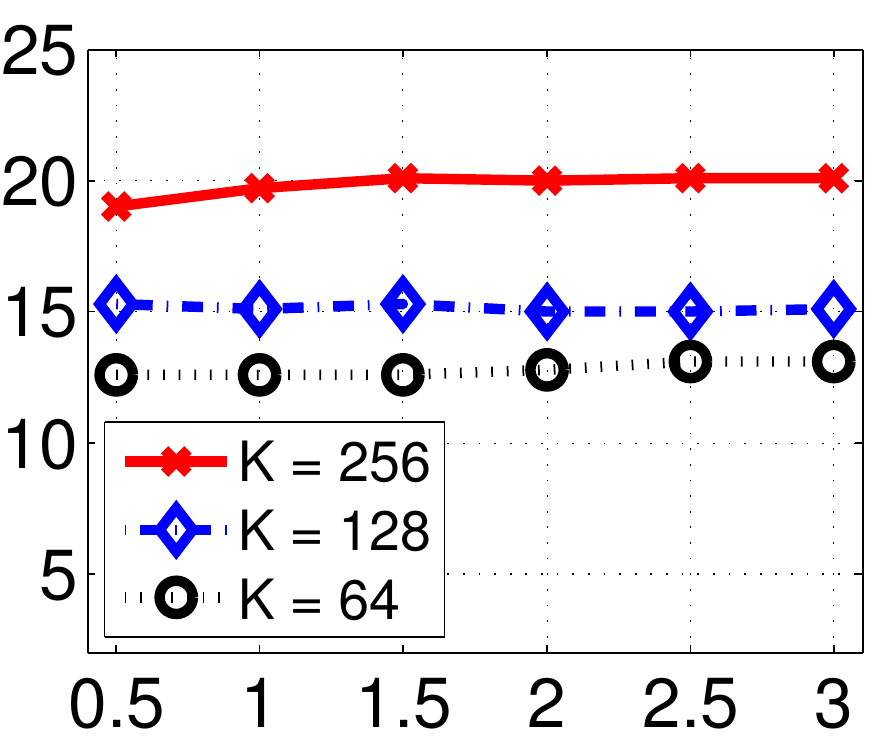}\\
    \vspace{-1ex}
    {\sf\footnotesize~~~segment length (sec)}
  \end{minipage}
\end{minipage}
\vspace{-1ex}
\caption{As per Fig.~\ref{fig:Influence_L_FV}, but showing the performance of the \mbox{\bf PI-BoW} variant (where the Fisher vector representation is replaced with BoW representation).}
\label{fig:Influence_L_BoW}
  
\end{figure}

We have evaluated the performance of two variants of the proposed system:
{\bf (1)} probabilistic integration with Fisher vectors (\mbox{\bf PI-FV}),
and
{\bf (2)} probabilistic integration with BoW histograms (\mbox{\bf PI-BoW}),
where the Fisher vector representation is replaced with BoW representation.
We start our experiments by studying the influence of the segment length $L$, expressed in terms of seconds.
The results are reported in Figs.~\ref{fig:Influence_L_FV} and~\ref{fig:Influence_L_BoW}, in terms of average accuracy over the folds.

Using the PI-FV variant (Fig.~\ref{fig:Influence_L_FV}), we found that using $L=1s$ and $K=256$ leads to the best performance on the \mbox{s-KTH dataset}.
For the \mbox{CMU-MMAC} dataset, the best performance is obtained with $L=2.5s$ and $K=64$. % 40.9\%, 
Note that using larger values of $K$ (ie., $128$ and $256$) leads to worse performance.
We attribute this to the large variability of appearance in the dataset,
where the training data may not be a good representative of the test data.
Consequently, using a large value of $K$ may lead to overfitting to the training data.

The optimal segment length for each dataset is different.
We attribute this to the \mbox{s-KTH} dataset containing short videos whose duration is between $1s$ and $7s$,
while CMU-MMAC has a large range of action durations between $0.1s$ and $108s$.
While the optimal values of $L$ and $K$ differ across the datasets,
the results also show that relatively good overall performance across both datasets
can be obtained with \mbox{$L=1s$} and \mbox{$K=64$}.

The results for the PI-BoW variant are shown in Fig.~\ref{fig:Influence_L_BoW}.
The best performance for the PI-BoW variant on the s-KTH dataset is obtained using $L=1s$ and $K=256$,
while on the \mbox{CMU-MMAC} dataset it is obtained with $L=2.5s$ and $K=256$.
These are the same values of $L$ and $K$ as for the PI-FV variant.
However, the performance of the PI-BoW variant is consistently worse than the PI-FV variant on both datasets.
This can be attributed to the better representation power of FV, as described in Section~\ref{sec:related}.
Note that the visual dictionary size $K$ for BoW is usually higher in order to achieve performance similar to FV.
However, due to the large size of the CMU-MMAC dataset, and for direct comparison purposes, we have used the same range of $K$ values throughout the experiments.

\subsection{Confusion Matrices}

\begin{figure}[!tb]
\centering
\begin{minipage}{1\textwidth}
\centering

\includegraphics[height=0.45\textheight]{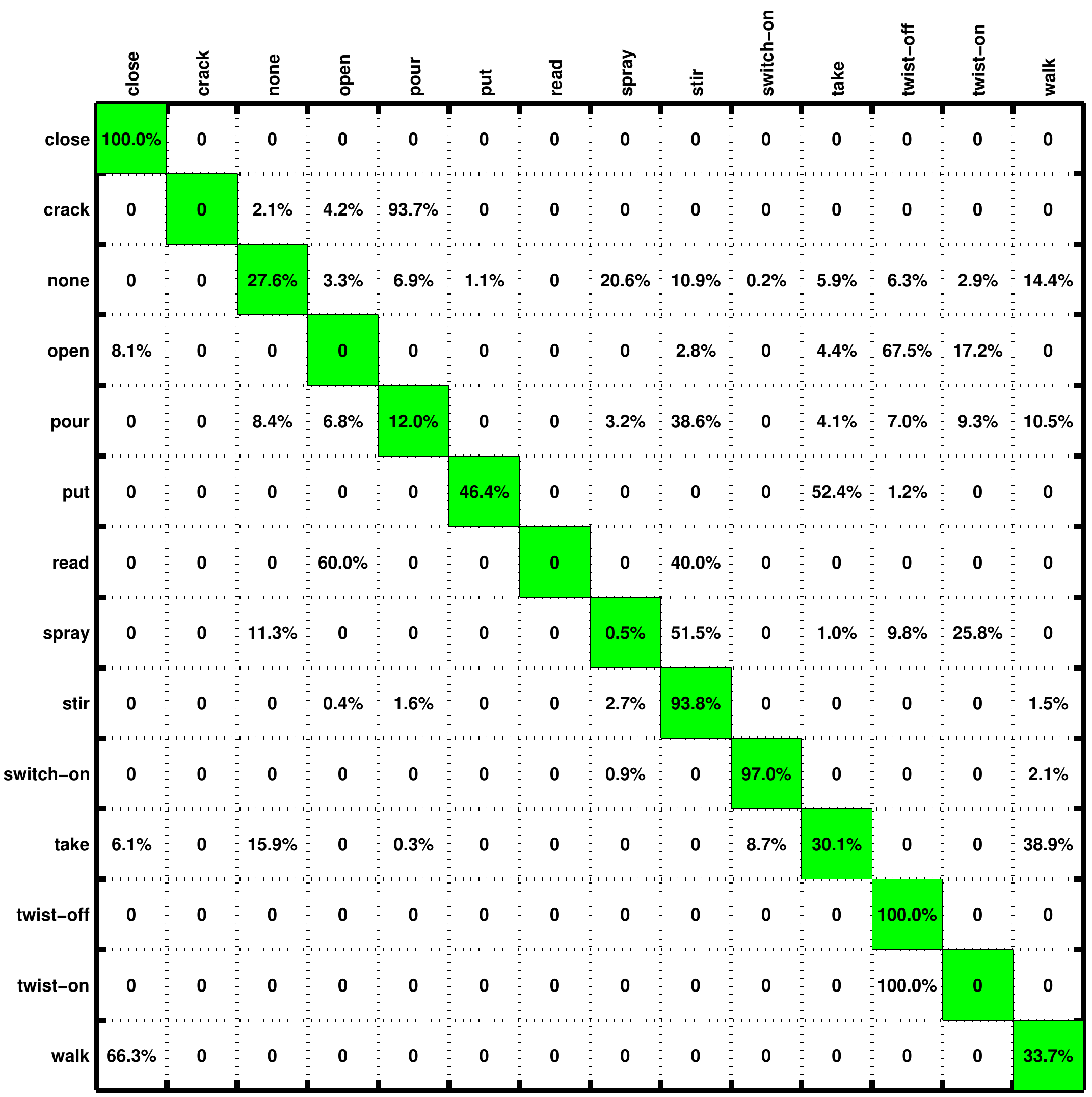}
\vspace{-2.5ex}
\caption{Confusion matrix for the PI-FV variant on the \mbox{CMU-MMAC} dataset.}
\label{fig:CF_CMU_FV}

\vspace{1ex}

\includegraphics[height=0.45\textheight]{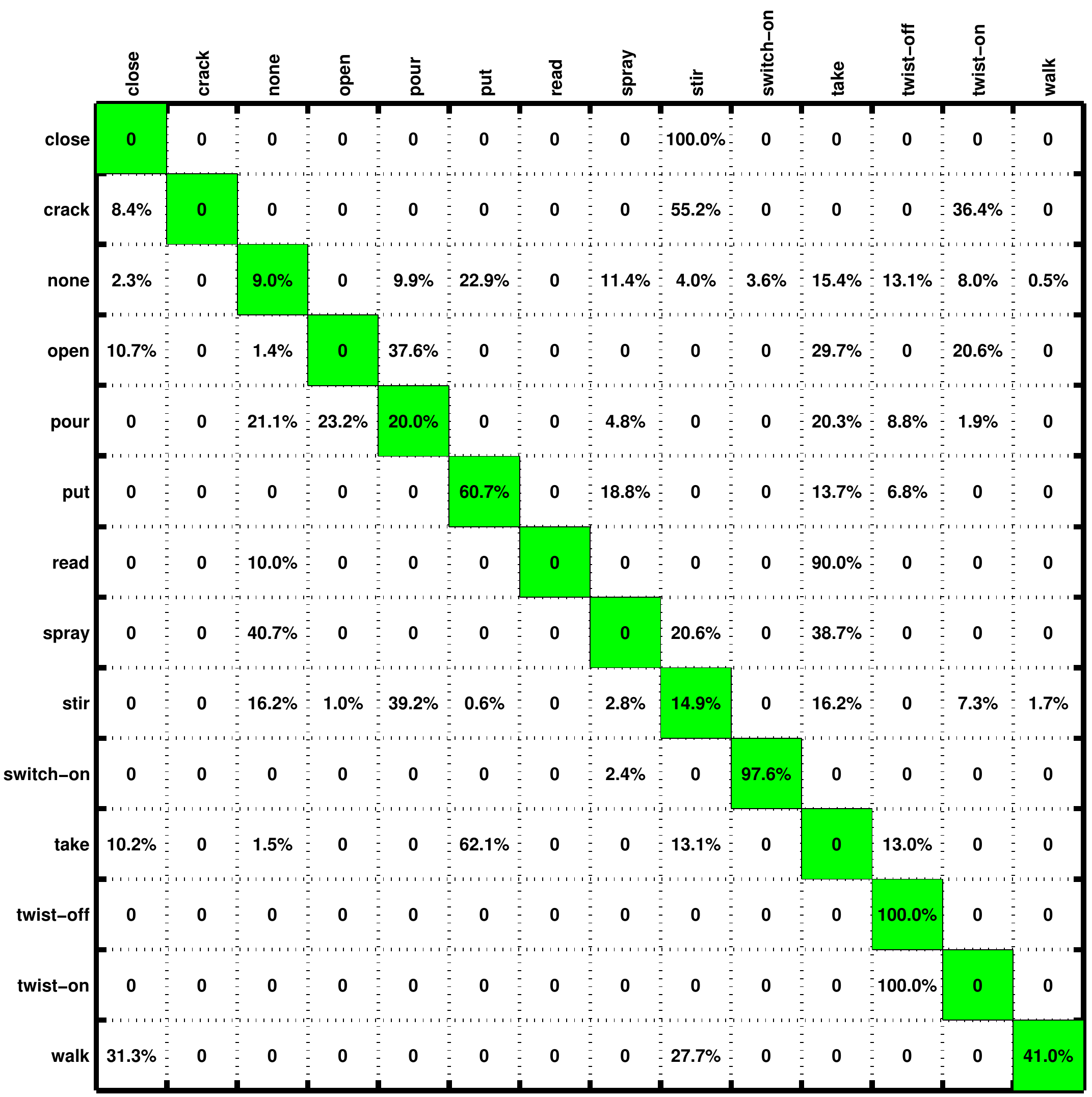}
\vspace{-2.5ex}
\caption{As per Fig.~\ref{fig:CF_CMU_FV}, but using the PI-BoW variant.}
\label{fig:CF_CMU_BoW}

\end{minipage}
\end{figure}

Figs.~\ref{fig:CF_CMU_FV} and~\ref{fig:CF_CMU_BoW} show the confusion matrices for the \mbox{PI-FV} and \mbox{PI-BoW} variants on the CMU-MMAC dataset.
The confusion matrices show that in 50\% of the cases (actions), the PI-BoW variant is unable to recognise the correct action.
Furthermore, the PI-FV variant on average obtains better action segmentation than PI-BoW.

For five actions ({\it crack}, {\it open}, {\it read}, {\it spray}, {\it twist-on}), \mbox{PI-FV} has accuracies of $0.5\%$ or lower.
Action {\it crack} implies crack and pour eggs into a bowl, but it's annotated only as {\it crack},
leading to confusion between {\it crack} and {\it pour}.
We suspect that actions {\it read} and {\it spray} are poorly modelled due to lack of training data; they are performed by a reduced number of subjects.
Action {\it twist-on} is confused with {\it twist-off} which are essentially the same action.

\clearpage
\subsection{Comparison with GMM and HMM-MIO}

We have compared the performance of the \mbox{PI-FV} and \mbox{PI-BoW} variants
against the \mbox{HMM-MIO}~\cite{Borzeshi2013} and stochastic modelling~\cite{Joha2014b} approaches previously used for multi-action recognition.
The comparative results are shown in Table~\ref{table:results}.

The proposed PI-FV method obtains the highest accuracy of 85.0\% and 40.9\% for the s-KTH and CMU-MMAC datasets, respectively. 
In addition to higher accuracy, the proposed method has other advantages over previous techniques.
There is just one global GMM (representing the visual vocabulary).
This is in contrast to \cite{Joha2014b} which uses one GMM (with a large number of components) for each action, leading to high computational complexity.
The HMM-MIO method in~\cite{Borzeshi2013} requires the search for many optimal parameters (as described in Section~\ref{subsec:prev_multi_action_recog}),
whereas the proposed method has just two parameters ($L$ and $K$).

\begin{table}[!tb]
\centering
\caption{Comparison of the proposed methods (PI-FV and \mbox{PI-BoW}) against several recent approaches on the stitched version of the KTH dataset \mbox{(s-KTH)} and the challenging \mbox{CMU-MMAC} dataset.}
\vspace{-2ex}
\begin{tabular}{c||c|c}
\textbf{Method} & \textbf{~~~~s-KTH~~~~} & \textbf{CMU-MMAC}  \\ \hline\hline
HMM-MIO~\cite{Borzeshi2013}           &  $71.2\%$  & $38.4\%$ \\ 
Stochastic Modelling~\cite{Joha2014b} &  $78.3\%$  & $33.7\%$ \\ 
PI-FV~~~                              &  $\bm{85.0}\%$ & $\bm{40.9}\%$\\ 
PI-BoW                                &  $48.0\%$  & $20.1\%$ \\ 
\end{tabular}
\label{table:results}
\end{table}

\subsection{Wall-Clock Time}

Lastly, we provide an analysis of the computational cost (in terms of wall-clock time) of our system and the stochastic modelling approach.
The wall-clock time is measured under optimal configuration for each system,
using a Linux machine with an Intel Core processor running at 2.83~GHz.
On the s-KTH dataset, the stochastic modelling system takes on average 228.4 minutes to segment and recognise a multi-action video.
In comparison, the proposed system takes 5.6 minutes, which is approximately 40 times faster.

\section{Conclusions and Future Work}
\label{sec:conclusions}

In this paper we have proposed a hierarchical approach to multi-action recognition that performs joint segmentation and classification in videos.
Videos are processed through overlapping temporal windows.
Each frame in a temporal window is represented using selective low-level spatio-temporal features
which efficiently capture relevant local dynamics and do not suffer from the instability and imprecision exhibited by STIP descriptors~\cite{Laptev2005}.
Features from each window are represented as a Fisher vector,
which captures the first and second order statistics.
Rather than directly classifying each Fisher vector,
it is converted into a vector of class probabilities.
The final classification decision for each frame (action label)
is then obtained by integrating the class probabilities at the frame level,
which exploits the overlapping of the temporal windows.
The proposed approach has a lower number of free parameters than previous methods which use dynamic programming or HMMs~\cite{Borzeshi2013}.
It is also considerably less computationally demanding compared to modelling each action directly with a GMM~\cite{Joha2014b}.

Experiments were done on two datasets:
s-KTH (a~stitched version of the KTH dataset to simulate multi-actions),
and the more challenging \mbox{CMU-MMAC} dataset (containing realistic multi-action videos of food preparation).
On s-KTH, the proposed approach achieves an accuracy of 85.0\%,
considerably outperforming two recent approaches based on GMMs and HMMs which obtained 78.3\% and 71.2\%, respectively.
On \mbox{CMU-MMAC}, the proposed approach achieves an accuracy of 40.9\%,
outperforming the GMM and HMM approaches which obtained 33.7\% and 38.4\%, respectively.
Furthermore, the proposed system is on average 40 times faster than the GMM based approach. 

Possible future areas of exploration include the use of the fast Fisher vector variant proposed in~\cite{Simonyan2013},
where for each sample the deviations for only one Gaussian are calculated.
This can deliver a large speed up in computation, at the cost of a small drop in accuracy~\cite{Parkhi14}.

~

\begin{footnotesize}

\noindent
{\bf Acknowledgements}. 
NICTA is funded by the Australian Government through the Department of Communications, as well as the Australian Research Council through the ICT Centre of Excellence program.
\end{footnotesize}

\renewcommand{\baselinestretch}{1.05}\small\normalsize

\bibliographystyle{ieee}
\bibliography{references}

\end{document}